\documentclass[pdflatex,sn-mathphys-num]{sn-jnl}

\usepackage{graphicx}%
\usepackage{multirow}%
\usepackage{amsmath,amssymb,amsfonts}%
\usepackage{amsthm}%
\usepackage{mathrsfs}%
\usepackage[title]{appendix}%
\usepackage{xcolor}%
\usepackage{textcomp}%
\usepackage{manyfoot}%
\usepackage{booktabs}%
\usepackage{algorithm}%
\usepackage{algorithmicx}%
\usepackage{algpseudocode}%
\usepackage{listings}%
\usepackage{makecell}%
\usepackage{subcaption}
\usepackage{booktabs}   
\usepackage{arydshln}   

\theoremstyle{thmstyleone}%
\theoremstyle{thmstyletwo}%

\theoremstyle{thmstylethree}%

\begin{document}

\title[Article Title]{From Passive to Proactive: A Hierarchical Multi-Agent Framework for Automated Medical Pre-Consultation}

\author[1]{\fnm{ChengZhang} \sur{Yu}}
\equalcont{These authors contributed equally to this work.}

\author[1]{\fnm{YingRu} \sur{He}}
\equalcont{These authors contributed equally to this work.}

\author[1]{\fnm{Hongyan} \sur{Cheng}}

\author[1]{\fnm{Nuo} \sur{Cheng}}

\author[2]{\fnm{Zhixing} \sur{Liu}}

\author[1]{\fnm{Dongxu} \sur{Mu}}

\author[1]{\fnm{Zhangrui} \sur{Shen}}

\author[1]{\fnm{Yang} \sur{Gao}}

\author*[1]{\fnm{Zhanpeng} \sur{Jin}}
\email{zjin@scut.edu.cn}

\affil[1]{\orgname{South China University of Technology}, \country{China}}

\affil[2]{\orgname{The Third Affiliated Hospital of Sun Yat-sen University}, \country{China}}

\abstract{

\textbf{Background:} Global healthcare systems face critical challenges from increasing patient volumes and limited consultation times, with primary care visits averaging under 5 minutes in many countries. Pre-consultation processes---encompassing triage and structured history-taking---offer potential solutions but remain limited by passive interaction paradigms and context management challenges in existing AI systems.

\textbf{Objective:} This study introduces a hierarchical multi-agent framework that transforms passive medical AI systems into proactive inquiry agents through autonomous task orchestration, addressing the limitations of current pre-consultation technologies.

\textbf{Methods:} We developed an eight-agent architecture with centralized control mechanisms for dynamic medical consultation. The framework decomposes pre-consultation into four primary tasks: Triage ($T_1$), History of Present Illness collection ($T_2$), Past History collection ($T_3$), and Chief Complaint generation ($T_4$), with $T_1$--$T_3$ further divided into 13 domain-specific subtasks. A Controller agent coordinates specialized agents through dynamic completion assessment, adaptive prompt generation, and hierarchical task management. The system was evaluated on 1,372 de-identified electronic health records a Chinese medical platform, with performance assessed across multiple foundation models (GPT-OSS 20B, Qwen3-8B, Phi4-14B). Clinical quality was independently rated by 18 licensed physicians using standardized criteria.

\textbf{Results:} The framework achieved 87.0\% accuracy for primary department triage and 80.5\% for secondary department classification. Task completion rates reached 98.2\% using agent-driven scheduling versus 93.1\% with sequential processing. Physician evaluation (n=18) yielded mean scores of 4.56 for Chief Complaints, 4.48 for History of Present Illness, and 4.69 for Past History on a 5-point scale, with low inter-rater variability ($\mathrm{SD} < 0.23$). The system demonstrated robust cross-model generalizability without task-specific fine-tuning.

\textbf{Conclusions:} The hierarchical multi-agent framework demonstrates feasibility in transforming passive pre-consultation systems into proactive, physician-like inquiry agents. The model-agnostic architecture supports flexible deployment while maintaining clinical quality standards. While validation on Chinese-language records limits immediate generalizability, these findings suggest a promising pathway for enhancing pre-consultation efficiency. Further prospective validation with real patient interactions is warranted.

}




\maketitle

\section{Introduction}
\label{sec1}

The global healthcare system faces a critical challenge: consultation times are often insufficient for comprehensive patient assessment. Systematic reviews across 67 countries reveal that primary care visits last 5 minutes or less in nations representing half the world's population \cite{irving2017international}, with some regions documenting consultations as brief as 4.3 minutes \cite{wang2022consultation}. These temporal constraints not only elevate physician burnout but fundamentally compromise diagnostic quality and patient safety \cite{malene2024aging, pooyan2018resource}. Pre-consultation—defined as the systematic process of guiding patients toward appropriate departments (triage) and collecting essential medical history prior to the formal encounter—has emerged as a vital workflow innovation to alleviate these systemic pressures \cite{zhakhina2023pre, DeGroot2022doc}.

Recent advances in large language models (LLMs) have catalyzed the development of conversational AI for automated patient information collection \cite{bickmore2015consent, tang2024medagents}. Early systems primarily utilized rule-based dialogue trees or retrieval-augmented mechanisms to improve response adaptability \cite{lewis2021retrieval}. More recently, the field has shifted toward collaborative multi-agent architectures to handle complex clinical tasks, with frameworks demonstrating capabilities in on-demand task decomposition \cite{Archiki2023adapt} and modular coordination among specialized agents \cite{wang2024tdag, wang2025consultation}. However, existing approaches exhibit a fundamental limitation: they predominantly employ \textit{passive interaction paradigms} that respond to patient inputs rather than proactively guiding structured inquiry. This reactive design fails to replicate physician-directed questioning—a cornerstone of diagnostic quality where clinicians systematically explore symptoms, assess severity, and rule out differential diagnoses through deliberate inquiry sequences \cite{gawlik2024historytaking}. Furthermore, even sophisticated multi-agent architectures struggle with extended medical dialogues, experiencing up to 39\% performance degradation due to context window limitations and the "loss-in-middle" phenomenon \cite{laban2025llms}—a critical deficiency given that comprehensive medical evaluations typically require 10–20 conversational exchanges.

We hypothesized that a hierarchical multi-agent framework with autonomous task orchestration, building upon decentralized coordination principles \cite{xiao2025agent}, could transform passive medical AI systems into proactive inquiry agents capable of physician-like structured questioning. To test this hypothesis, we pursued three primary objectives: (i) to design and implement a hierarchical multi-agent framework incorporating dynamic subtask completion assessment and adaptive prompt generation for pre-consultation; (ii) to formalize the pre-consultation workflow into structured primary tasks—Triage ($T_1$), History of Present Illness (HPI) collection ($T_2$), Past History (PH) collection ($T_3$), and Chief Complaint (CC) generation ($T_4$)—with domain-specific subtask decomposition; and (iii) to evaluate the framework's performance in triage accuracy, task completion.

To achieve these objectives, we developed a hierarchical multi-agent framework coordinating eight specialized agents through a central Controller. The framework introduces three core innovations addressing the identified limitations: (1) \textbf{dynamic subtask completion assessment} enabling a Monitor agent to evaluate information gathering progress across 13 predefined medical domains using clinically-informed criteria; (2) \textbf{adaptive prompt generation} where a Prompter agent synthesizes patient responses with accumulated context to formulate targeted, proactive follow-up questions; and (3) \textbf{hierarchical task management} where a Controller agent balances macroscopic diagnostic progression with microscopic symptom detail collection through priority-based orchestration. This architecture maintains contextual coherence across extended dialogues while ensuring an assertive inquiry style. We evaluated the framework using 1,372 validated medical records, assessing triage accuracy, cross-model generalizability, and physician-rated documentation quality to ensure clinical utility.

\section{Methods}
\label{sec:methods}

This study presents the development and evaluation of a hierarchical multi-agent framework for automated pre-consultation in healthcare settings. The methodology encompasses dataset construction, system architecture design, and evaluation procedures.

\subsection{Study Setting and Dataset Construction}
\label{subsec:dataset}

\subsubsection{Data Sources}
We constructed a comprehensive dataset for medical triage classification by collecting Electronic Health Records (EHRs) from iiyi.com\footnote{\url{https://bingli.iiyi.com/}}, a well-established public medical platform with over 20 years of operational history in China. This platform serves as a large-scale online forum for sharing de-identified electronic medical records contributed by users from diverse hospitals across different regions. The platform maintains strict privacy standards by applying proper de-identification processes to all shared EHR data, ensuring complete protection of patient privacy information.

Using systematic web crawling techniques, we initially collected 8,172 electronic health records from the platform. To ensure temporal relevance and reflect current medical practices, all collected records were sourced exclusively from the past five years, guaranteeing that our dataset captures contemporary medical terminology, treatment approaches, and diagnostic patterns.

\subsubsection{Data Processing and Quality Assurance}
We implemented a rigorous two-stage data processing pipeline to ensure high-quality annotations and eliminate inconsistencies.

The first stage involved comprehensive quality filtering where we systematically examined all essential fields including History of Present Illness (HPI), Past History (PH), and Chief Complaint (CC). Records were excluded if they met any of the following criteria: (1) empty or missing fields, (2) obvious data entry errors, (3) abnormally short HPI descriptions, or (4) excessive similarity between HPI and PH content indicating potential data duplication. This filtering process reduced our initial collection from 8,172 to 2,263 records.

The second stage employed a hybrid validation approach combining automated prediction with expert medical review. We first applied our triage prediction model to each remaining record using HPI, PH, and CC information as input features. Records where the predicted department assignment matched the original ground truth were automatically validated and retained. For cases where automated prediction disagreed with the original triage decision, qualified medical professionals conducted thorough manual verification to ensure accuracy and reliability of ground truth labels.

\begin{figure}[h]
\centering
\includegraphics[width=0.6\textwidth]{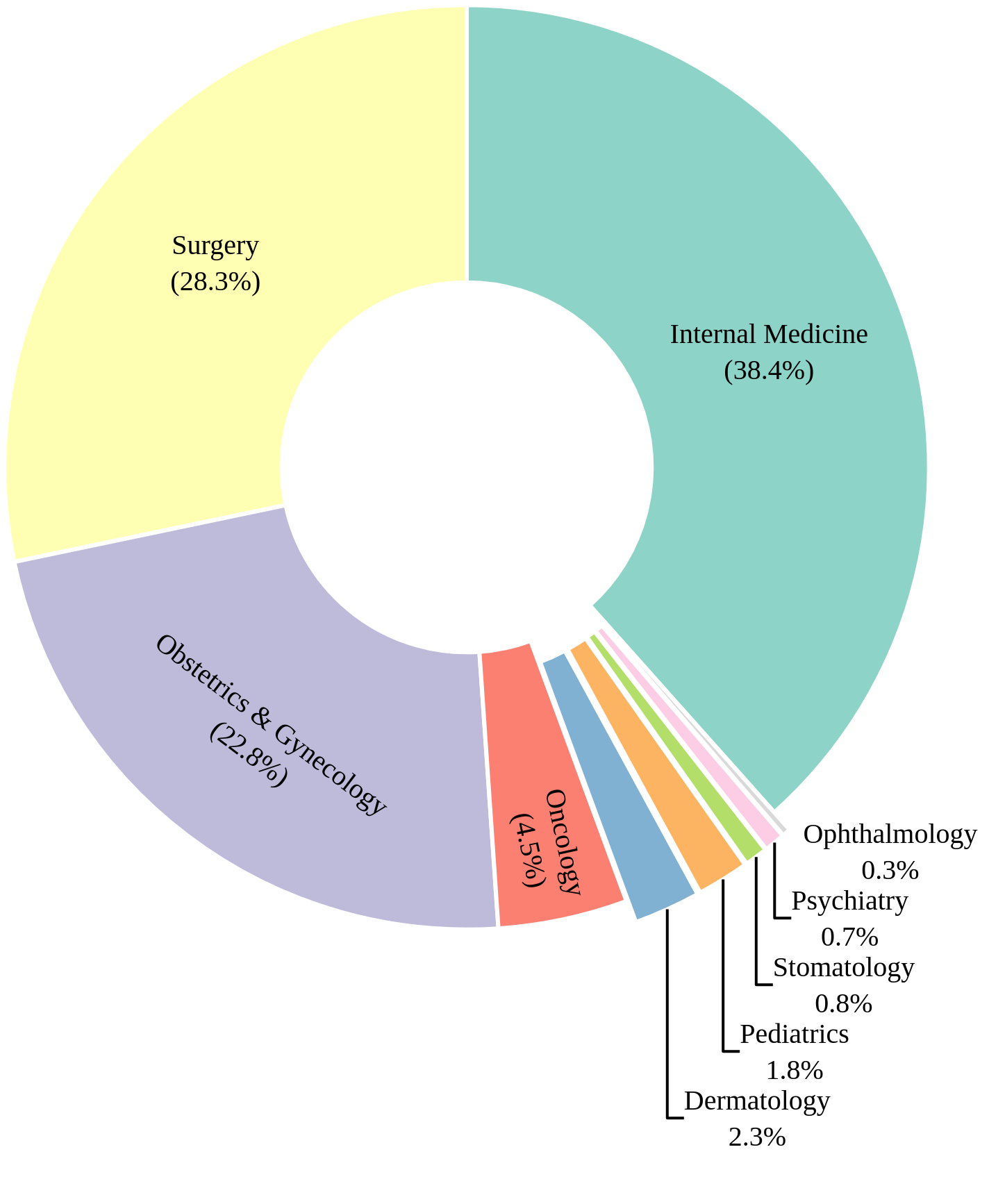}  
\caption{Distribution of clinical departments in our dataset. The chart shows the proportional representation of primary medical specialties, with Internal Medicine comprising the largest share (38.4\%) and Ophthalmology the smallest (0.3\%).}
\label{fig:primary_distribution}
\end{figure}

Our final curated dataset comprises 1,372 validated medical records organized in a hierarchical structure reflecting real-world medical triage systems. The dataset encompasses 9 primary medical departments and 35 secondary sub-specialties, providing comprehensive coverage for both coarse-grained and fine-grained classification tasks. Figure~\ref{fig:primary_distribution} illustrates the distribution of cases across primary medical departments.

\subsection{The Hierarchical Multi-Agent Framework}
\label{subsec:framework}

\subsubsection{Task Formalization}
We formalize the pre-consultation process $T$ as comprising two stages: Triage and History Taking. Given its large volume of information, heterogeneity of content, and the requirement for structuring doctor–patient communication, History Taking is further organized into three modules: Chief Complaint (CC), History of Present Illness (HPI), and Past History (PH) \cite{gawlik2024historytaking}.

The pre-consultation process is modeled as a hierarchical optimization structure comprising four main task groups:
\begin{align}
    T &= \{T_1, T_2, T_3, T_4\} \\
    T_1 &= \{t_{11}, t_{12}\} \\
    T_2 &= \{t_{21}, t_{22}, \ldots, t_{26}\} \\
    T_3 &= \{t_{31}, t_{32}, \ldots, t_{35}\}
\end{align}
where $T_1$ represents Triage, $T_2$ represents HPI collection, $T_3$ represents PH collection, and $T_4$ represents CC generation. This hierarchical modeling approach facilitates a balance between global optimality and efficient completion of individual tasks.

\begin{table}[!ht]
\centering
\caption{Task definitions in the pre-consultation framework}
\label{tab:task_definitions}
\small
\renewcommand{\arraystretch}{1.3}

\begin{tabular}{
>{\raggedright\arraybackslash}p{0.15\textwidth}
>{\raggedright\arraybackslash}p{0.30\textwidth}
>{\raggedright\arraybackslash}p{0.45\textwidth}
}
\toprule[1.5pt]
\textbf{Task} & \textbf{Subtasks} & \textbf{Description} \\
\midrule[1pt]

\multirow{2}{*}{Triage} 
    & Primary Department Identification 
    & Determine the primary department the patient should visit. \\
    & Secondary Department Identification 
    & Identify the specific secondary department based on the primary department. \\[0.3em]

\hdashline[0.5pt/2pt]

\multirow{6}{*}{HPI Collection} 
    & Onset 
    & Record the time, location, mode of onset, prodromal symptoms, and possible causes or triggers. \\
    & Main Symptom Characteristics 
    & Describe the location, nature, duration, severity, and aggravating/relieving factors of main symptoms in chronological order. \\
    & Disease Progression 
    & Describe the progression and evolution of the illness in chronological order. \\
    & Accompanying Symptoms 
    & Record accompanying symptoms and their relationship with the main symptoms. \\
    & Diagnostic and Therapeutic History 
    & Record whether the patient has undergone examinations or treatments after onset, and their outcomes if applicable. \\
    & General Condition 
    & Briefly record the patient's mental state, sleep, appetite, bowel and bladder functions, and body weight after onset. \\[0.3em]

\hdashline[0.5pt/2pt]

\multirow{5}{*}{PH Collection} 
    & Disease History 
    & Record the patient's history of past illnesses, including infectious diseases such as tuberculosis and hepatitis. \\
    & Immunization History 
    & Inquire about the patient's vaccination history. \\
    & Surgical and Trauma History 
    & Record the patient's history of surgeries and traumas. \\
    & Blood Transfusion History 
    & Inquire about the patient's history of blood transfusions and any adverse reactions. \\
    & Allergy History 
    & Inquire about the patient's history of food or drug allergies. \\

\bottomrule[1.5pt]
\end{tabular}
\end{table}

Based on existing clinical guidelines \cite{gawlik2024historytaking} and discussions with multiple physicians, we decomposed the first three tasks into domain-specific subtasks (Table~\ref{tab:task_definitions}). Unlike the decomposable collection tasks, CC generation ($T_4$) is a comprehensive integrative task that operates directly on the incremental recording process, refining information through continuous accumulation of dialogue content and dynamic evolution of the HPI.

\subsubsection{Multi-Agent Architecture}

\begin{figure}[!ht]
\centering
\includegraphics[width=1\columnwidth]{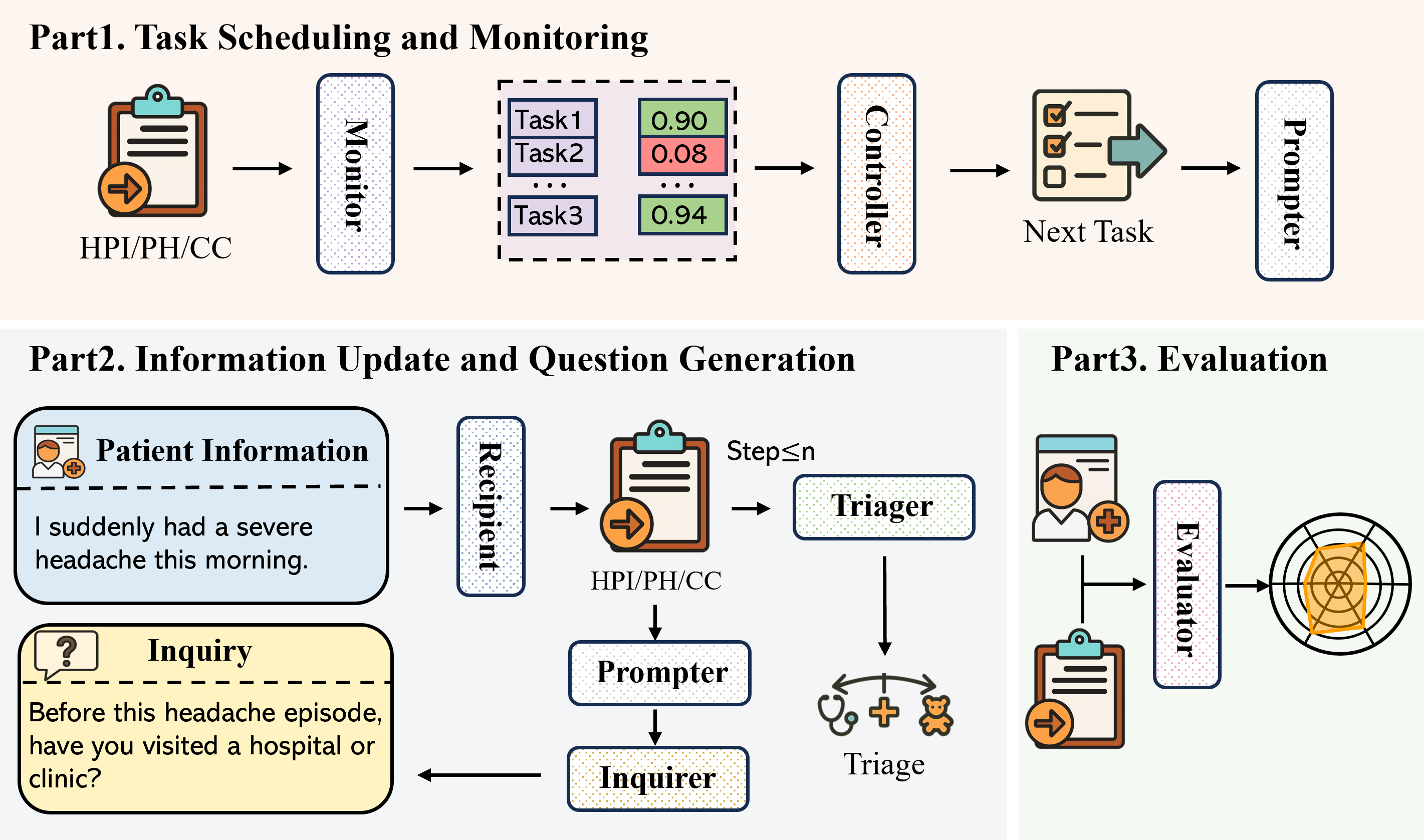} 
\caption{Hierarchical multi-agent framework architecture for medical consultation workflow. Parts 1-3 correspond to Sec. \ref{Task Scheduling and Monitoring}, \ref{Information Update and Question Generation} respectively.}
\label{fig1}
\end{figure}

The multi-agent framework addresses the challenge that pre-consultation tasks involve both decomposable, structured collection components and context-dependent, integrative generation components, which are difficult to handle simultaneously with a single model. Through collaborative mechanisms, different tasks are handled by independent autonomous entities while achieving global coordination during interactions. The overall architecture is illustrated in Figure~\ref{fig1}.

\begin{table}[!ht]
\centering
\caption{Overview of the Agents and Their Functions}
\label{tab:overview of the agents}
\small
\begin{tabular}{p{0.2\textwidth}p{0.7\textwidth}}
\toprule
\textbf{Agent Name} & \textbf{Function} \\
\midrule
Virtual Patient & Generate realistic clinical presentations based on validated medical records \\
Recipient & Update and maintain evolving medical records throughout dialogue sessions\\ 
Triager & Perform hierarchical department-level triage with primary and secondary classification \\
Monitor & Assess subtask completion status using clinical validity and completeness metrics \\
Controller & Select optimal next subtask through priority-based scheduling algorithm\\
Prompter & Formulate context-aware inquiry strategies based on current task objectives\\
Inquirer & Produce clinically appropriate medical questions maintaining dialogue coherence\\
Evaluator & Provide comprehensive performance assessment across multiple quality dimensions\\
\bottomrule
\end{tabular}
\end{table}

We designed eight core agents covering global scheduling, question–answer generation, and task evaluation. This configuration balances task coverage and system complexity: the number of agents is sufficient to allow fine-grained task handling while avoiding redundancy and information fragmentation. Each agent has a clearly defined functional role (Table~\ref{tab:overview of the agents}).

\subsubsection{Task Scheduling and Monitoring}
\label{Task Scheduling and Monitoring}
When the system receives the $k$-th round of updated patient information (including Triage, HPI, PH, and CC), the current task group for evaluation is selected based on the completion order of different sections. If the current task group remains consistent with the previous round, the system inherits the pending task set $T_P$ from the previous round.

The \textit{Monitor} agent evaluates each subtask within the pending task set $T_P$ by incorporating newly acquired patient information. Evaluation is based on two dimensions: clinical semantic validity and task completeness, with scores ranging from 0 to 1. Subtasks achieving scores above the threshold of 0.85 are removed from the pending task set, while those scoring below remain for further processing:
\begin{align}
    T_P^{k+1} = \{t_{ij} \mid S(t_{ij}) < 0.85, t_{ij} \in T_P^{k}\}
\end{align}
where $S(\cdot)$ denotes the evaluation function of the Monitor. The threshold mechanism balances information integrity and dialogue efficiency: an excessively high threshold may lead to redundant follow-up questions, while a threshold that is too low risks missing critical clinical details.

Once the Monitor completes evaluation, the \textit{Controller} agent performs global scheduling based on real-time pending task states. It integrates task priority, dialogue context, and patient clinical information to select the highest-priority subtask as the current task $T_C$:
\begin{align}
    T_C^k = C(T_P^k)
\end{align}
The Controller then generates corresponding inquiry instructions passed to the Prompter and Inquirer agents.

\subsubsection{Information Update and Question Generation}
\label{Information Update and Question Generation}
Task scheduling alone is insufficient to support fully interactive consultation. The system must dynamically update patient information during dialogue and generate targeted clinical questions to ensure professionality and coherence of inquiry.

Upon receiving natural language expressions from the patient, the \textit{Recipient} agent integrates new information into the existing medical record and incrementally updates CC, HPI, and PH:
\begin{align}
    (\mathrm{HPI}_{k+1}, \mathrm{PH}_{k+1}, \mathrm{CC}_{k+1}) = F(\mathrm{HPI}_k, \mathrm{PH}_k, \mathrm{CC}_k, \mathcal{C}_{k+1})
\end{align}
where $\mathcal{C}_{k+1}$ is the complete dialogue up to turn $k+1$, and $F$ denotes the Recipient's update function. This mechanism allows patient history to evolve across dialogue turns rather than remain at the level of isolated statements.

The \textit{Prompter} agent generates guiding prompts for targeted questioning by integrating updated HPI, PH, and CC from the Recipient with the current task $T_C$ from the Controller. This ensures generated prompts align precisely with current task objectives while remaining within the scope of clinician–patient interactions.

Guided by prompts from the Prompter and the updated medical record, the \textit{Inquirer} agent produces context-sensitive clinical questions reflecting scheduling objectives while maintaining logical continuity and medical professionalism.

Additionally, when task $T_1$ (Triage) has not been completed, the \textit{Triager} agent analyzes symptom characteristics and provides triage recommendations by constructing contextualized clinical interpretations based on structured taxonomies. Once $T_1$ is completed, the Triager is no longer invoked.

\subsubsection{Safety Constraints in Agent Design}
\label{Safety Constraints in Agent Design}
To mitigate potential risks associated with LLM-generated content in medical contexts, we implemented prompt-level safety constraints across all agents. Specifically, the Recipient and Inquirer agents were explicitly instructed 
to (1) refrain from fabricating or inferring medical information not directly stated by the patient, (2) avoid leading or suggestive questioning that could bias patient responses, and (3) acknowledge uncertainty when information is incomplete rather than generating plausible but unverified content. These constraints were reinforced through few-shot examples demonstrating both appropriate and inappropriate agent behaviors within the system prompts. However, we acknowledge that this approach relies on prompt-based guidance rather than architectural enforcement mechanisms.

\subsection{Framework Evaluation Methods}
\label{subsec:evaluation}

The performance of the proposed framework was assessed using both quantitative metrics and qualitative evaluation dimensions.

\subsubsection{Quantitative Assessment: Interaction and Content Quality}
Within the multi-agent framework, the \textit{Evaluator} is designed as an independent assessment agent responsible for systematically measuring system performance. The Evaluator does not participate in actual clinician–patient dialogue and does not influence task scheduling or subtask selection; its function is entirely limited to post hoc evaluation.

To ensure comprehensive and rigorous evaluation, the Evaluator operates on two complementary levels reflecting the dual objectives of clinical pre-consultation:

\paragraph{Interaction-Level Assessment}
At the interaction level, the Evaluator examines: (1) whether system questions follow clinical logic and task objectives, (2) clarity and professionalism of language, (3) completeness of information collection, and (4) overall professionalism demonstrated during inquiry.

\paragraph{Content-Level Assessment}
At the content level, the Evaluator assesses alignment of the three key medical record elements—Chief Complaint, History of Present Illness, and Past History—against reference standards to ensure consistency and fidelity.

\begin{table}[!ht]
\centering
\caption{Evaluation Dimensions for the Evaluator}
\label{tab:evaluator_dimensions}
\small
\begin{tabular}{p{0.25\textwidth}p{0.65\textwidth}}
\toprule
\textbf{Dimension} & \textbf{Description} \\
\midrule
Clinical Inquiry (CI) & Assess completeness, professionalism, and logical flow of medical history collection process \\
Communication Quality (CQ) & Evaluate fluency, clarity, and empathy in doctor-patient dialogue interactions \\
Information Completeness (IC) & Measure the integrity, systematicity, and focus of information collection across all domains \\
Overall Professionalism (OP) & Assess domain knowledge accuracy, appropriate terminology usage, and structured clinical reasoning \\
CC Similarity (CCS) & Compare generated chief complaint content and semantics with reference standard \\
HPI Similarity (HPIS) & Measure coverage and structural alignment compared to reference History of Present Illness \\
PH Similarity (PHS) & Evaluate correspondence with reference Past History including completeness and accuracy \\
\bottomrule
\end{tabular}
\end{table}

Detailed definitions and criteria for each dimension are provided in Table~\ref{tab:evaluator_dimensions}.

\subsubsection{Scoring Methodology}

\begin{table}[!ht]
\centering
\caption{Scoring Criteria for Past History Similarity Assessment}
\label{tab:ph_similarity_criteria}
\small
\begin{tabular}{cp{0.8\textwidth}}
\toprule
\textbf{Score} & \textbf{Description} \\
\midrule
0 & Past History not mentioned or contains insufficient information for meaningful assessment \\
1 & Partially accurate; loosely related to true PH but contains notable omissions or inaccuracies \\
3 & Generally accurate; covers most key elements of true PH with minor deviations or missing details \\
5 & Highly accurate; closely matches true PH with comprehensive coverage and precise medical terminology \\
\bottomrule
\end{tabular}
\end{table}

The Evaluator uses a 0–5 scale for all seven dimensions, where 0 represents severe omission or error and 5 represents ideal clinical level. Complementarily, qualitative criteria interpret each score range with specific explanations and examples. For instance, Table~\ref{tab:ph_similarity_criteria} presents scoring standards for Past History Similarity (PHS), considering information coverage, chronological consistency, and accuracy of medical terminology. This combination of quantitative and qualitative assessments ensures objective scoring while providing interpretable feedback for system optimization.

\section{Results}
\label{sec:results}

The performance of the proposed hierarchical multi-agent framework was evaluated using quantitative metrics assessing triage accuracy, task completion, consultation quality, and scheduling strategy effectiveness, complemented by qualitative feedback from licensed physicians. Evaluation was conducted on 1,372 validated medical records from our curated dataset.

\subsection{Triage Performance}
\label{subsec:triage_results}

The framework demonstrated progressive improvement in medical triage accuracy through iterative refinement. Table~\ref{tab:department_classification} presents the classification accuracy across multiple iteration steps. Primary department classification accuracy increased from 83.0\% in the initial step to 87.0\% by step 4, representing a 4.0 percentage point improvement. Similarly, secondary department classification showed substantial gains, rising from 75.4\% to 80.5\%, an improvement of 5.1 percentage points.

\begin{table}[!ht]
\centering
\caption{Performance progression across iteration steps for primary and secondary department classification.}
\label{tab:department_classification}
\begin{tabular}{ccc}
\hline
\textbf{Iteration Step} & \textbf{Primary Department} & \textbf{Secondary Department} \\ \hline
Step 1 & 83.0\% & 75.4\% \\
Step 2 & 85.6\% & 78.7\% \\
Step 3 & 86.4\% & 80.2\% \\
Step 4 & 87.0\% & 80.5\% \\ \hline
\end{tabular}
\end{table}

\begin{figure}[!ht]
	\centering
	\includegraphics[width=1.05\columnwidth]{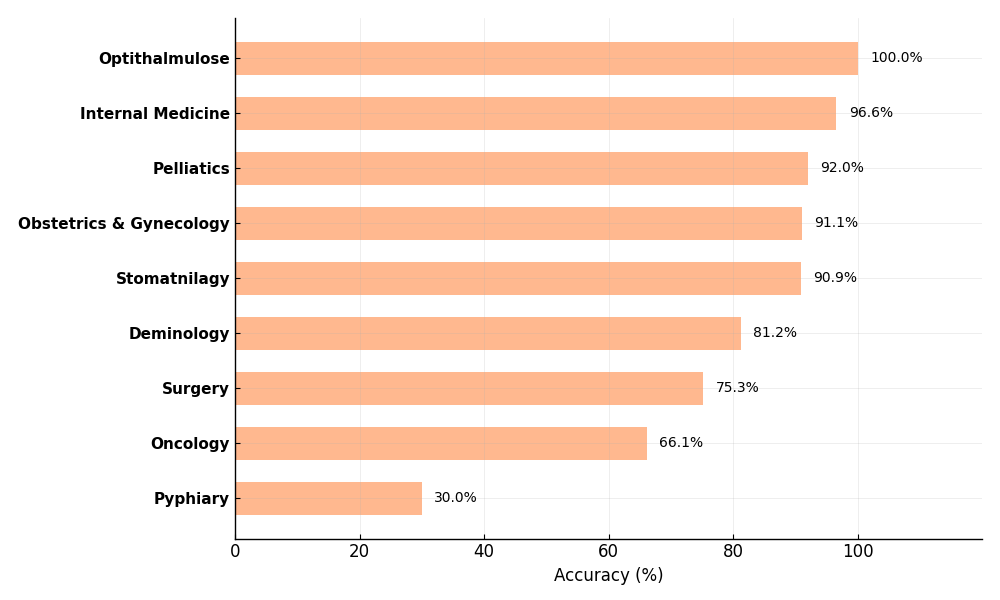}
	\caption{Triage accuracy across different medical departments. The bar chart shows significant variation in classification performance across specialties.}
\label{fig:department_distribution}
\end{figure}

Figure~\ref{fig:department_distribution} illustrates the triage accuracy distribution across different medical departments. Significant variability in performance was observed across specialties, with accuracy ranging from approximately 65\% to 95\%. Ophthalmology achieved the highest classification accuracy at 94.8\%, likely due to the distinctive nature of ocular symptoms that facilitate clear differentiation. In contrast, Psychiatry exhibited the lowest accuracy at 65.2\%. Further analysis revealed that misclassified psychiatry cases were predominantly routed to the neurology department, reflecting the inherent overlap between neurological and psychiatric presentations and potential variations in departmental structures across different hospital systems.

\subsection{Task Completion Across Foundation Models}
\label{subsec:task_completion}

To evaluate the generalizability of the hierarchical multi-agent architecture, we conducted zero-shot tests on three LLMs of different scales: GPT-OSS 20B, Qwen3-8B, and Phi4-14B. All models were deployed on a single NVIDIA A100 GPU with 60 parallel threads for concurrent processing. The maximum number of dialogue turns was set to 30 rounds per consultation session; samples that failed to complete all three task groups ($T_1$, $T_2$, and $T_3$) within this limit were classified as failures.

\begin{figure*}[b]
    \centering

    \begin{subfigure}[b]{0.45\textwidth}
        \centering
        \includegraphics[width=\linewidth]{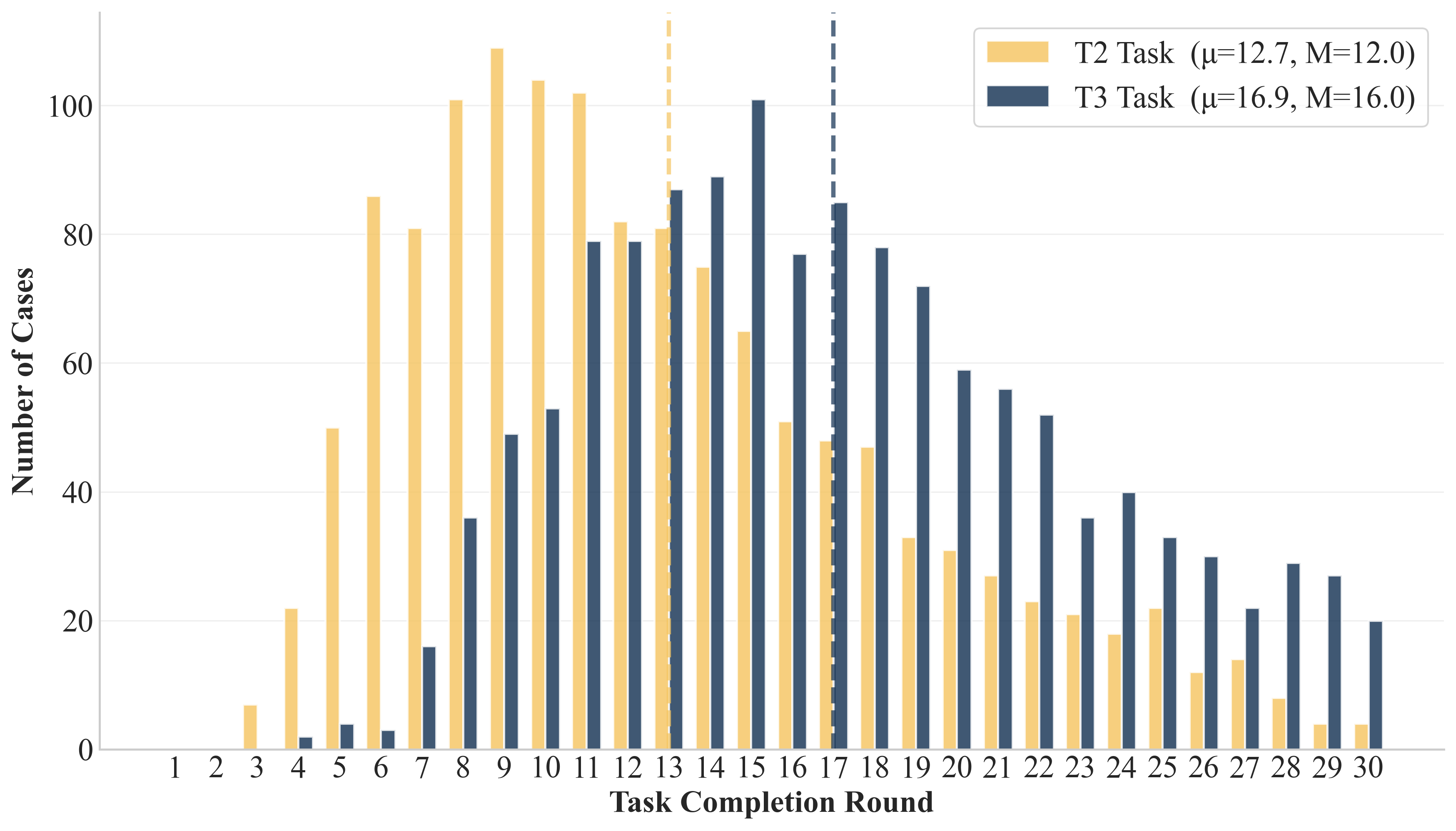} 
        \caption{GPT-OSS} 
        \label{fig:gpt-oss}
    \end{subfigure}
    \hfill 
    \begin{subfigure}[b]{0.45\textwidth}
        \centering
        \includegraphics[width=\linewidth]{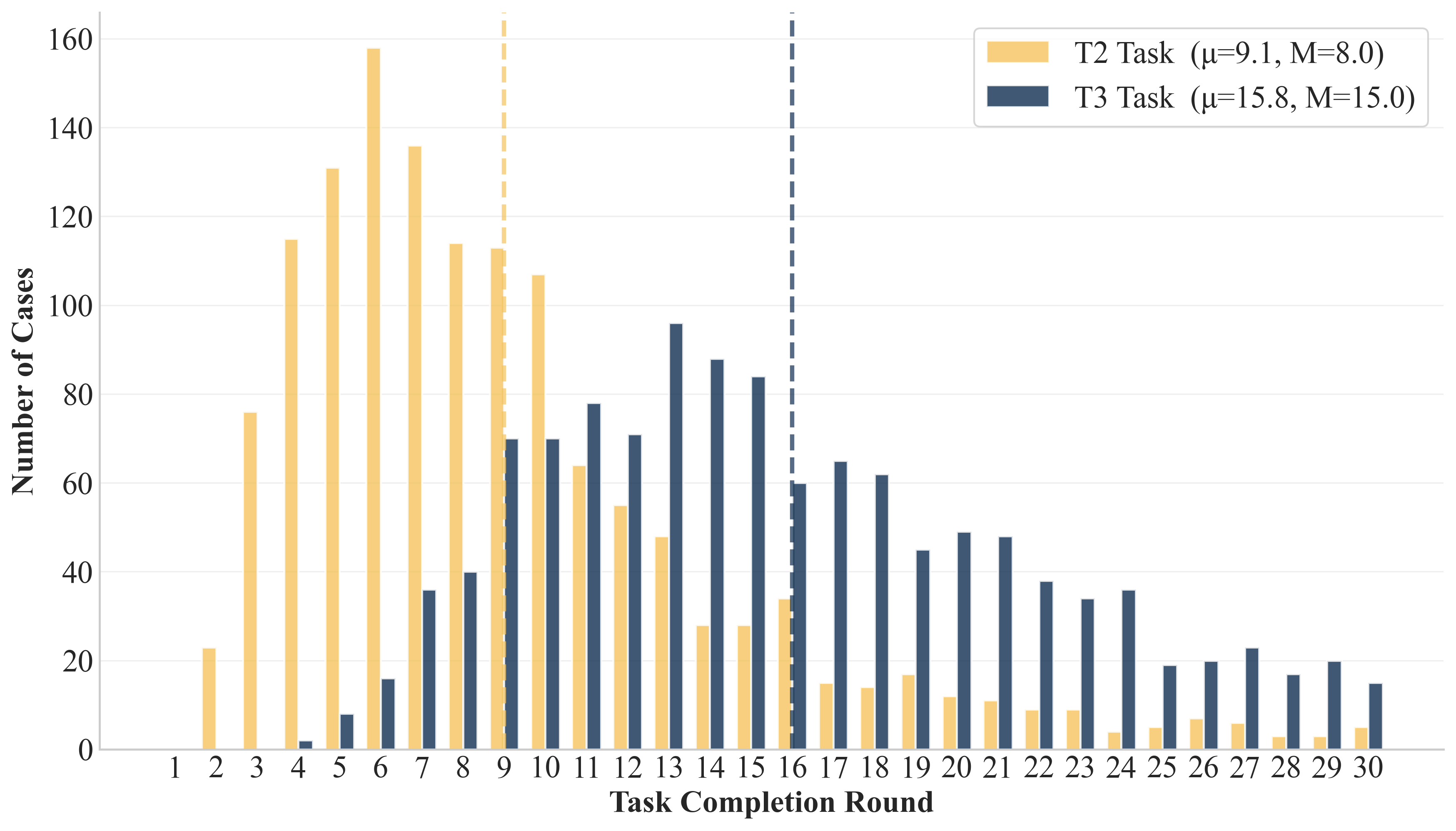} 
        \caption{Qwen3}
        \label{fig:qwen3}
    \end{subfigure}

    
    \begin{subfigure}[b]{0.45\textwidth}
        \centering
        \includegraphics[width=\linewidth]{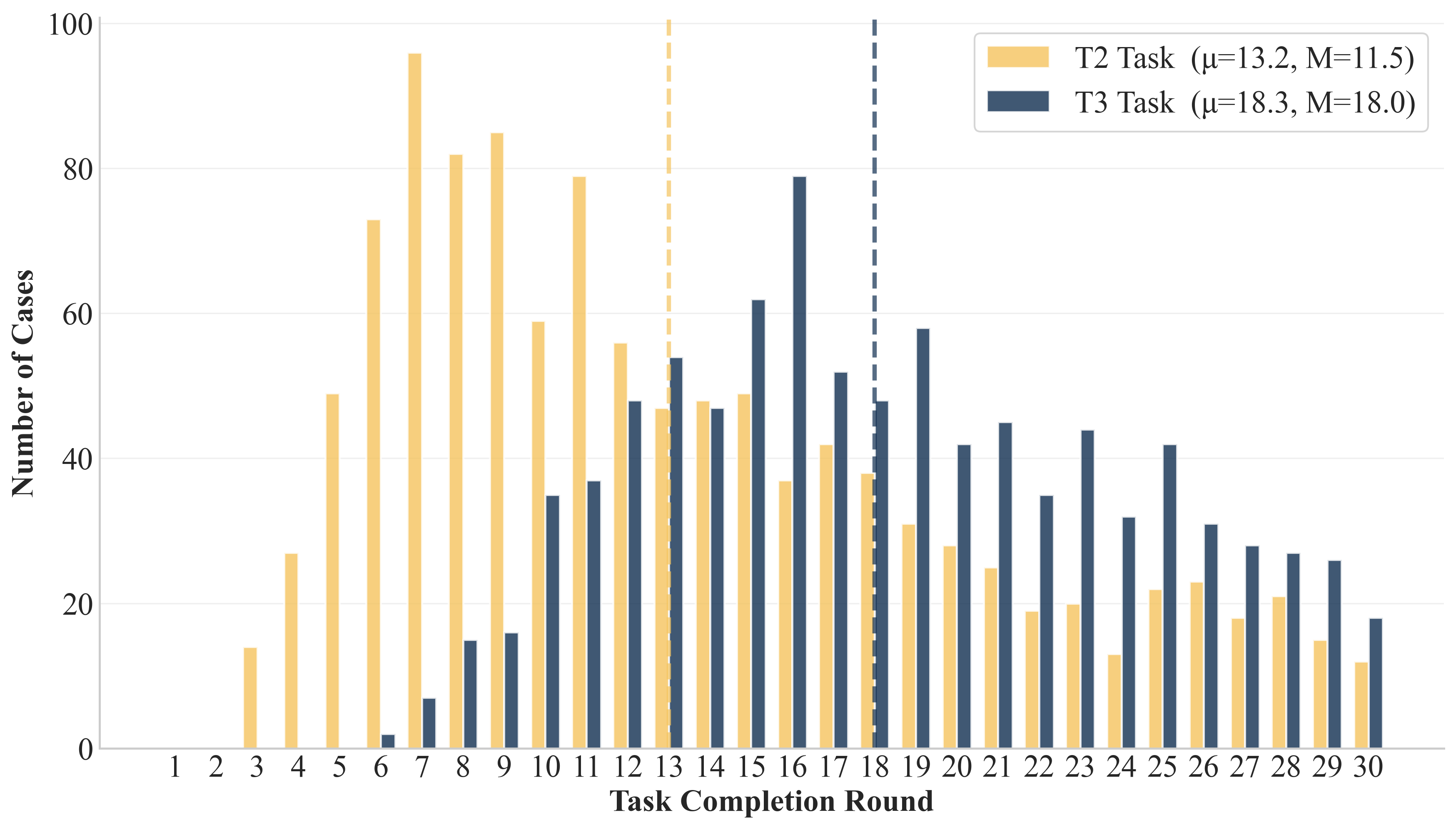} 
        \caption{Phi4}
        \label{fig:phi4}
    \end{subfigure}%
    \hfill 
    \begin{subfigure}[b]{0.45\textwidth}
        \centering
        \includegraphics[width=\linewidth]{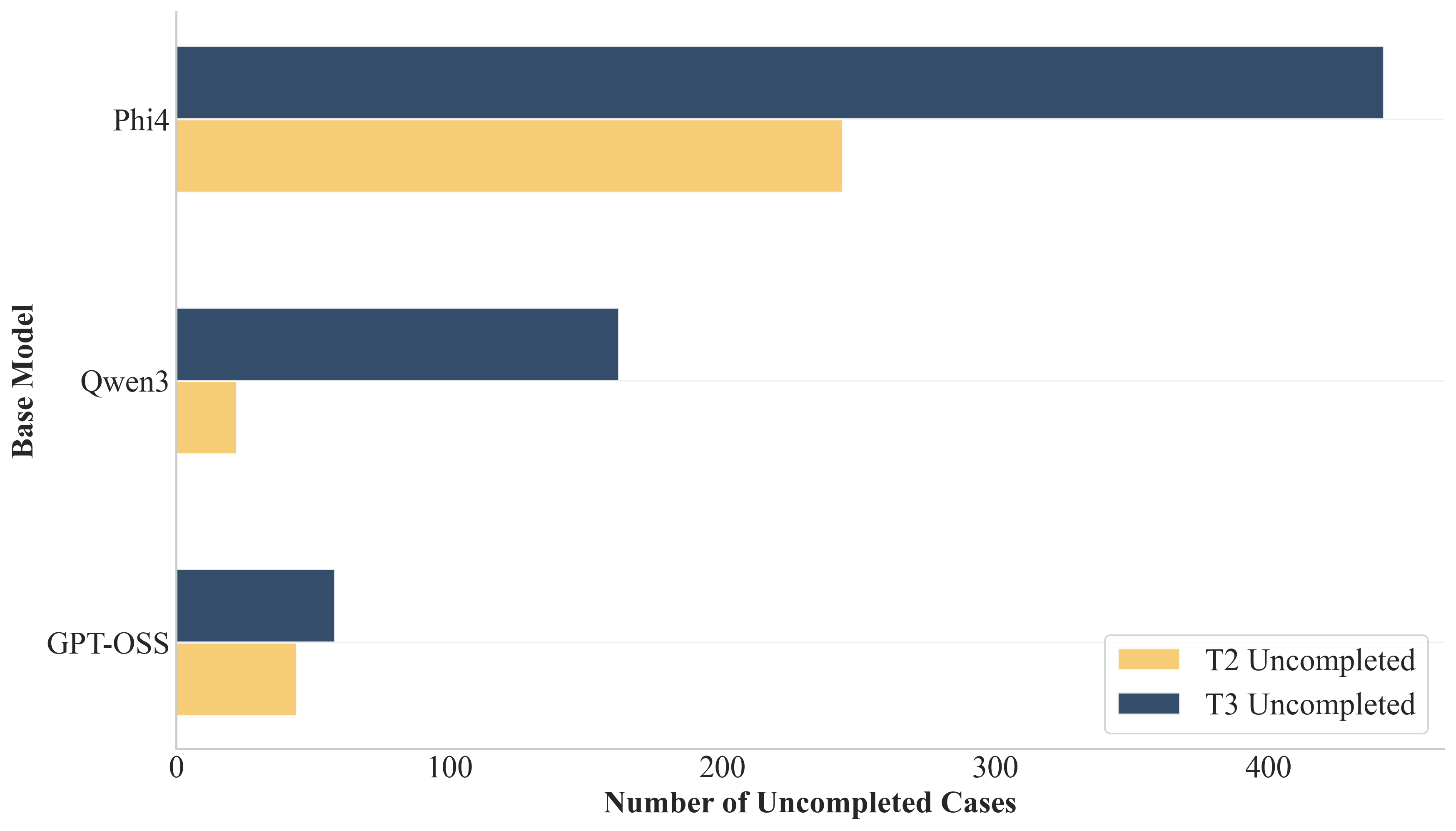}
        \caption{Uncompleted cases}
        \label{fig:uncompleted}
    \end{subfigure}

    \caption{Performance comparison across different foundation models. Subfigures (a)-(c) show results for GPT-OSS 20B, Qwen3-8B, and Phi4-14B respectively. Subfigure (d) illustrates the distribution of incomplete samples across all three models.}
    \label{fig:news} 
\end{figure*}

The results (Figure~\ref{fig:news}) demonstrate that our framework successfully guided all models to complete multi-turn medical consultations without any task-specific fine-tuning, highlighting its model-agnostic nature. GPT-OSS 20B achieved the most stable and reliable performance, with the lowest number of unfinished cases. The combined unfinished count for $T_2$ and $T_3$ tasks by GPT-OSS 20B was only 55.43\% of that achieved by Qwen3-8B and merely 8.08\% of Phi4-14B's unfinished cases.

Notably, Qwen3-8B demonstrated remarkable efficiency in task execution, completing both $T_2$ and $T_3$ tasks with fewer average rounds compared to GPT-OSS 20B. This efficiency advantage may be attributed to Qwen3-8B's stronger Chinese language capabilities, which are particularly relevant for our medical consultation scenarios. However, its lower completion rate compared to GPT-OSS 20B suggests that superior reasoning abilities remain crucial for ensuring task success. For Phi4-14B, overall performance lagged behind both GPT-OSS 20B and Qwen3-8B, which may relate to its earlier release date. Despite this limitation, the framework exhibited strong robustness, successfully guiding Phi4-14B to complete a considerable number of cases.

All models required more rounds to complete the more complex $T_3$ tasks, consistent with expectations and validating the soundness of our evaluation design. These results demonstrate that task completion success rate does not solely depend on model parameter size but rather correlates more strongly with language-specific capabilities and general reasoning abilities.

\subsection{Consultation Quality Evaluation}
\label{subsec:quality_evaluation}

\begin{figure}[!ht]
    \centering
    \includegraphics[width=1\linewidth]{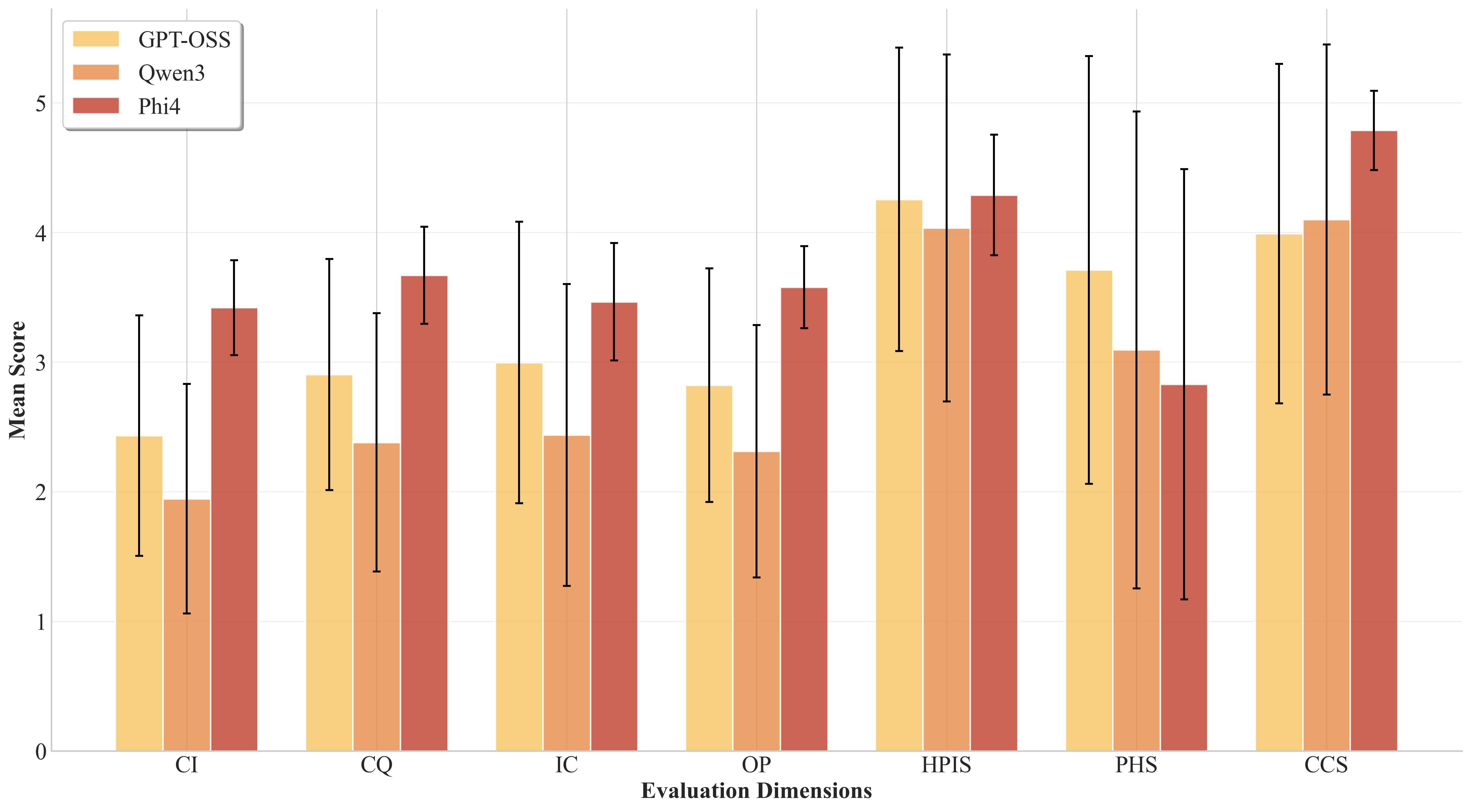}
    \caption{Mean scores and variance across seven evaluation dimensions for GPT-OSS 20B, Qwen3-8B, and Phi4-14B. Error bars indicate standard deviation.}
    \label{fig:bars}
\end{figure}

All three models achieved commendable evaluation scores across seven distinct dimensions (Figure~\ref{fig:bars}). To ensure consistency and fairness, the Evaluator component uniformly employed GPT-OSS 20B across all experiments.

In the three metrics reflecting information extraction effectiveness—Chief Complaint Similarity (CCS), Present Illness Similarity (HPIS), and Past History Similarity (PHS)—all models achieved average scores of at least 3.74 on a 5-point scale. Specifically, GPT-OSS 20B achieved an average of 3.99, Qwen3-8B scored 3.74, and Phi4-14B reached 3.99. These consistently high scores demonstrate that our system achieves excellent final outcomes across different models regardless of their individual characteristics.

Regarding the four metrics reflecting dialogue process quality—Clinical Inquiry (CI), Communication Quality (CQ), Information Completeness (IC), and Overall Professionalism (OP)—Phi4-14B demonstrated slightly stronger performance compared to GPT-OSS 20B and Qwen3-8B. This performance variation may be attributed to differences in linguistic styles across models, suggesting that certain conversational characteristics inherent to Phi4-14B align well with the dialogue evaluation criteria.

\subsection{Scheduling Strategy Comparison}
\label{subsec:scheduling_comparison}

We compared two representative scheduling strategies: Default Order (baseline sequential prompting following a predefined sequence designed with reference to clinical importance) and Agent Driven (autonomous coordination through intelligent agent selection).

\subsubsection{Task Completion Rate}
\begin{figure*}[!ht]
    \centering
    \begin{subfigure}[b]{0.48\textwidth}
        \centering
        \includegraphics[width=\linewidth]{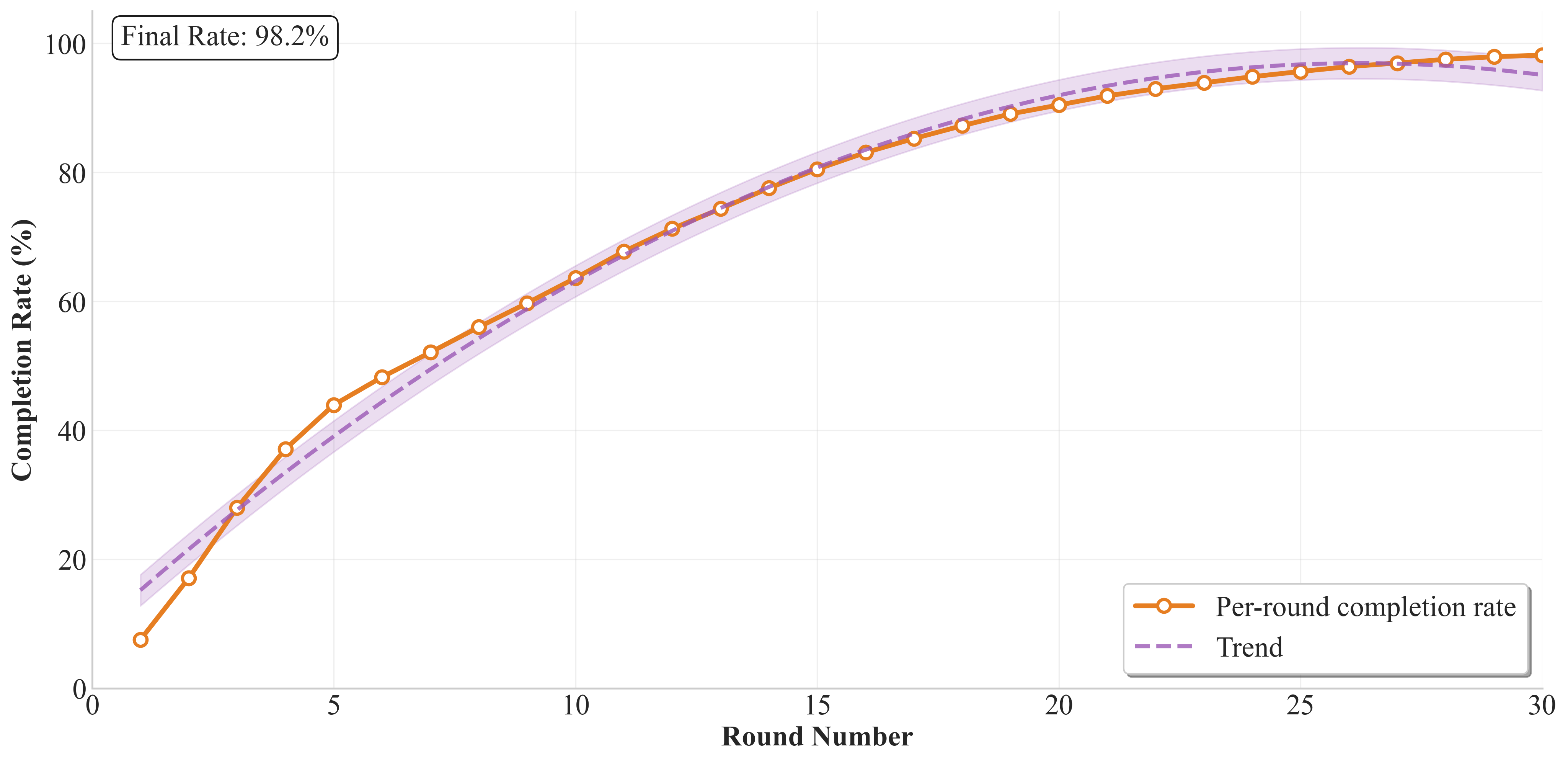} 
        \caption{Agent Driven} 
        \label{fig:agent_driven}
    \end{subfigure}%
    \hfill
    \begin{subfigure}[b]{0.48\textwidth}
        \centering
        \includegraphics[width=\linewidth]{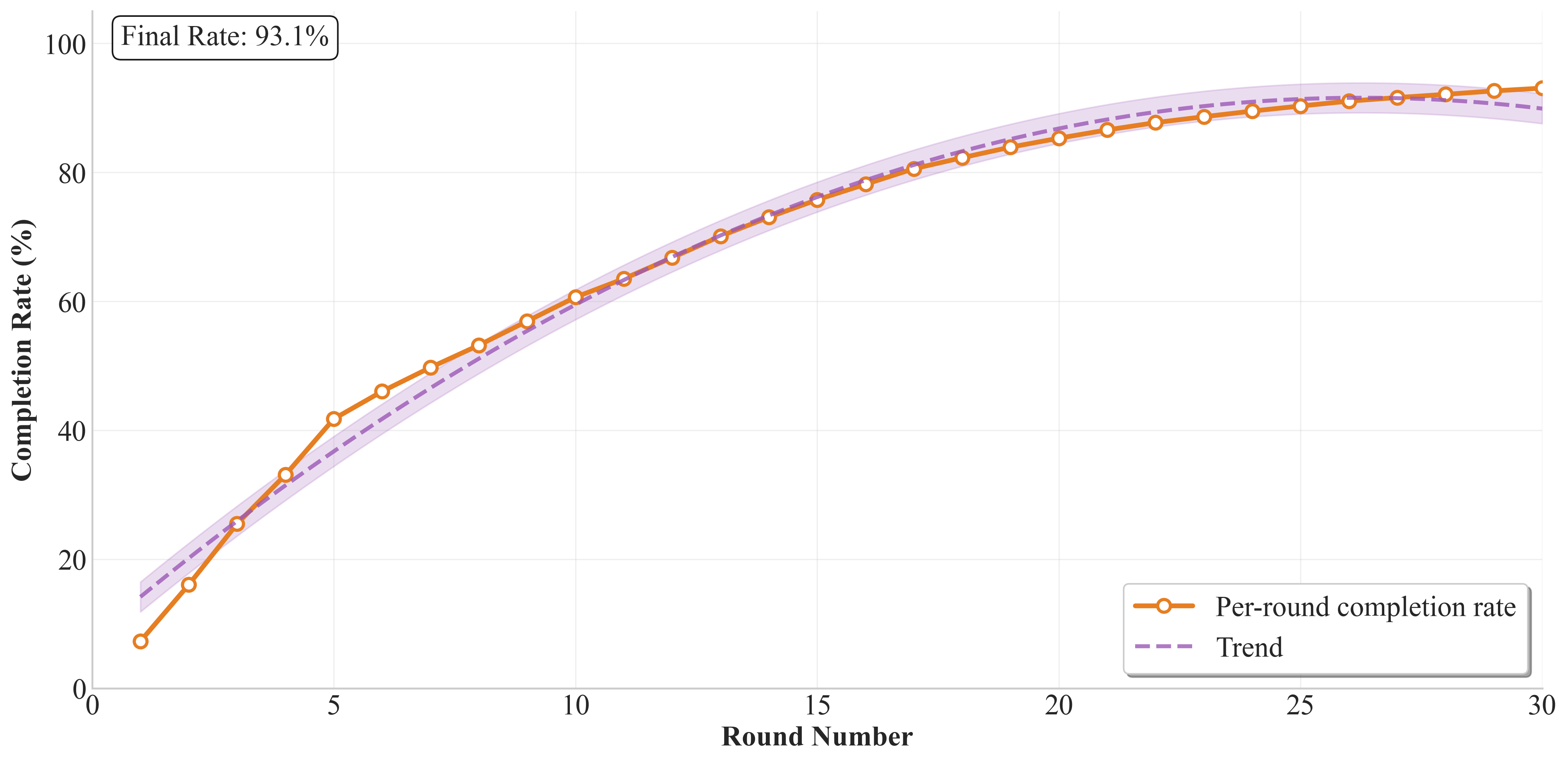} 
        \caption{Default Order}
        \label{fig:default_order}
    \end{subfigure}
\caption{Task completion rates under different scheduling strategies. The Agent Driven strategy achieves 98.2\% completion rate compared to 93.1\% for the Default Order approach.}
\label{fig:com-con}
\end{figure*}

As shown in Figure~\ref{fig:com-con}, both strategies exhibited monotonic upward trends in task completion across the 13 subtasks but differed in convergence levels and growth rates. The Agent Driven strategy demonstrated the fastest growth in the early phase and stabilized at the highest completion rate of 98.2\% around round 25. In contrast, the Default Order strategy maintained a steadier growth trajectory during the middle phase (rounds 5–15) but plateaued at a lower final completion rate of 93.1\%.

\subsubsection{Quality Assessment}
\begin{figure}[!ht]
    \centering
    \includegraphics[width=1\columnwidth]{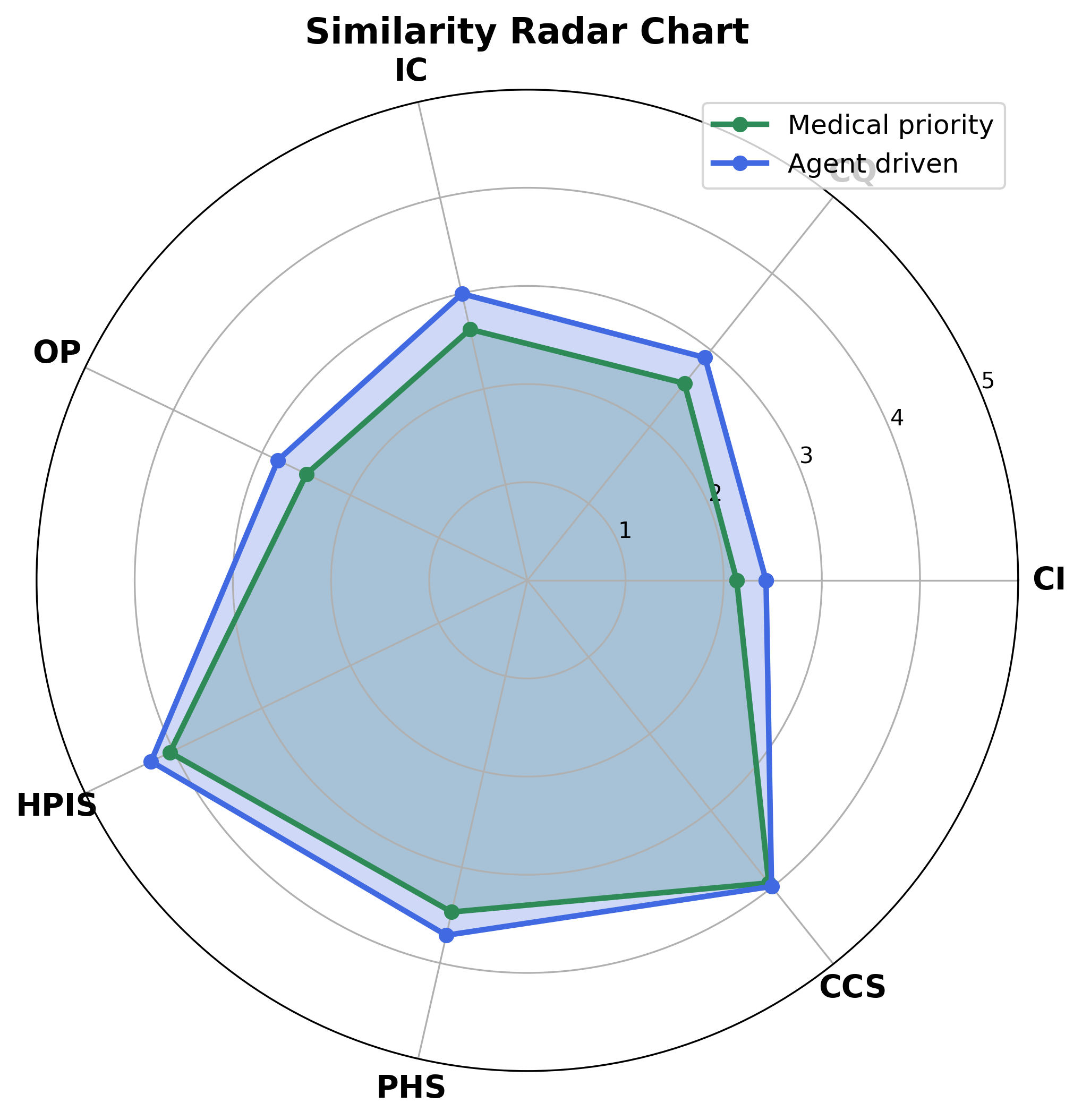}
    \caption{Evaluation scores across different scheduling strategies. Agent Driven demonstrates superior performance across most dimensions.}
    \label{fig:placeholder}
\end{figure}

Figure~\ref{fig:placeholder} presents the comparison of evaluation scores across scheduling strategies. The Agent Driven strategy consistently outperformed Default Order across most dimensions, achieving notable gains in IC and OP (reflecting stronger coherence and broader coverage) and showing clear advantages in PHS and HPIS (suggesting more complete and accurate medical history collection). While the two strategies achieved comparable performance in CCS, Agent Driven maintained a slight edge. The only area where Default Order remained competitive was CI, though its overall balance lagged behind.

\subsection{Physician Evaluation}
\label{subsec:physician_evaluation}

Although the Evaluator-based scoring offers consistency and reference value, its evaluation focus may differ from real-world pre-consultation settings. To validate clinical relevance, we incorporated expert physician evaluation as a complementary assessment.

A total of 40 samples were randomly selected from the dataset of 1,372 records. For each sample, 5 out of 20 dialogue turns were randomly sampled to present to evaluators, balancing comprehensive evaluation with maintaining physician focus and reducing cognitive load. The sampled dialogue turns, along with the final-round generated Chief Complaint (CC), History of Present Illness (HPI), and Past History (PH), were presented to 18 licensed physicians. Each physician independently assessed 20 samples, resulting in 9 physician evaluations per sample on a discrete scale from 1 to 5.

Table~\ref{tab:real_world_evaluation} presents the expert physician evaluation results. All average scores exceeded 4.45, indicating generally high quality, while relatively low standard deviations (below 0.23) demonstrated consistent agreement among raters.

\begin{table}[!ht]
\centering
\caption{Physician evaluation scores for generated medical documentation.}
\label{tab:real_world_evaluation}
\begin{tabular}{lcc}
\toprule
\textbf{Field} & \textbf{Mean Score} & \textbf{Standard Deviation} \\
\midrule
Chief Complaint & 4.56 & 0.22 \\
History of Present Illness & 4.48 & 0.13 \\
Past History & 4.69 & 0.10 \\
\bottomrule
\end{tabular}
\end{table}

\subsection{Representative Case Analysis}
\label{subsec:case_analysis}

To illustrate the framework's practical performance, we present a representative consultation case with corresponding physician evaluations.

\subsubsection{Dialogue Excerpt}
The following excerpt demonstrates the system's proactive inquiry capability:

\begin{quote}
\textbf{Patient:} My neck and shoulder have been hurting for half a month. It's hard to move.

\textbf{System:} Please think back carefully. When the pain started, did anything specific happen?

\textbf{Patient:} That night, I got up too quickly while turning over, and suddenly my right neck cramped with sharp pain. After resting for 10 minutes, it eased.

\textbf{System:} Can you describe the nature of the pain? Is it a dull constant ache, intermittent stabbing, or something else?

\textbf{Patient:} It's mostly spasmodic pain in the neck and right shoulder. Very intense during attacks, but subsides after rest. The left shoulder feels sore and stiff in the morning.

\textbf{System:} Any other symptoms, such as numbness in the arm, dizziness, or vision changes?

\textbf{Patient:} No, just the neck and shoulder pain.

\textbf{System:} Have you ever been diagnosed with any chronic diseases, like hypertension, diabetes, heart disease, hepatitis, etc.?

\textbf{Patient:} No, I've always been healthy.
\end{quote}

\subsubsection{Generated Documentation}
Based on the dialogue, the system generated the following medical documentation:

\textbf{Chief Complaint:} Neck and shoulder pain with limited mobility for half a month.

\textbf{History of Present Illness:} Patient developed neck and shoulder pain with restricted mobility half a month ago. One day ago, a sudden movement while getting up led to acute cramping pain in the right neck, which resolved after 10 minutes of rest. The pain is intense during attacks and alleviates with rest. Left shoulder shows noticeable morning soreness and limited mobility. Denies numbness or dizziness.

\textbf{Past History:} Denies any history of chronic diseases such as hypertension or diabetes. No prior surgeries. No known drug or food allergies.

\subsubsection{Case-Specific Physician Evaluation}
Nine medical practitioners evaluated this case, with results shown in Table~\ref{tab:case_evaluation}. The high scores and unanimous agreement on Past History (5.00) demonstrate the system's capability to generate clinically accurate and comprehensive documentation.

\begin{table}[!ht]
\centering
\caption{Physician evaluation scores for the representative case.}
\label{tab:case_evaluation}
\begin{tabular}{lccc}
\toprule
\textbf{Field} & \textbf{4 pt} & \textbf{5 pt} & \textbf{Mean Score} \\
\midrule
Chief Complaint & 2 & 7 & 4.78 \\
History of Present Illness & 3 & 6 & 4.67 \\
Past History & 0 & 9 & 5.00 \\
\bottomrule
\end{tabular}
\end{table}

\section{Discussion}
\label{sec:discussion}

\subsection{Summary of Main Findings}
\label{subsec:summary}

This study introduced and evaluated a novel hierarchical multi-agent framework designed to transform passive medical AI systems into proactive inquiry agents for pre-consultation. Our findings demonstrate the feasibility and effectiveness of this approach across multiple dimensions.

The framework achieved progressive improvement in triage accuracy, with primary department classification increasing from 83.0\% to 87.0\% and secondary department classification improving from 75.4\% to 80.5\% through iterative refinement (Table~\ref{tab:department_classification}). The multi-agent architecture demonstrated strong generalizability across three foundation models of different scales (GPT-OSS 20B, Qwen3-8B, and Phi4-14B), successfully guiding all models to complete multi-turn medical consultations without task-specific fine-tuning. Furthermore, the Agent Driven scheduling strategy achieved a 98.2\% task completion rate, substantially outperforming the Default Order baseline (93.1\%). Critically, physician evaluation of generated medical documentation yielded high scores across all fields (CC: 4.56, HPI: 4.48, PH: 4.69 on a 5-point scale) with low inter-rater variability, suggesting clinical acceptability of the framework's outputs.

The primary contribution of this work lies in proposing and validating a hierarchical task decomposition approach combined with intelligent agent coordination, offering a pathway to address the well-documented challenges of consultation time constraints and pre-consultation workflow inefficiencies in healthcare systems worldwide \cite{irving2017international,wang2022consultation}.

\subsection{Interpreting Framework Performance}
\label{subsec:interpreting}

The quantitative results regarding triage accuracy and task completion underscore the framework's potential to support clinical pre-consultation workflows. The consistent improvement in triage accuracy across iteration steps (Table~\ref{tab:department_classification}) validates our multi-agent approach, where agents collaboratively refine their decisions through structured dialogue and knowledge sharing. The 4.0 and 5.1 percentage point improvements in primary and secondary department classification, respectively, demonstrate that iterative refinement through agent coordination yields tangible performance gains.

The observed variability in triage performance across medical departments (Figure~\ref{fig:department_distribution}) warrants discussion. Ophthalmology achieved the highest accuracy (94.8\%), likely due to the distinctive nature of ocular symptoms that facilitate clear differentiation. In contrast, Psychiatry exhibited the lowest accuracy (65.2\%), with misclassified cases predominantly routed to the neurology department. This pattern reflects the inherent clinical overlap between neurological and psychiatric presentations—a challenge well-recognized in medical practice—and suggests that our framework appropriately captures the ambiguity inherent in certain clinical scenarios rather than artificially forcing definitive classifications.

The cross-model evaluation results (Figure~\ref{fig:news}) reveal important insights about framework generalizability. Despite substantial differences in model architectures and parameter scales, all three tested models successfully completed multi-turn consultations under the framework's guidance. Notably, task completion success did not solely depend on model parameter size: Qwen3-8B demonstrated superior efficiency (fewer rounds to completion) compared to the larger GPT-OSS 20B, potentially attributable to stronger Chinese language capabilities relevant to our consultation scenarios. This finding suggests that the hierarchical multi-agent architecture provides sufficient structure to elicit effective performance even from smaller models, while language-specific capabilities and reasoning abilities remain important factors influencing task success.

The comparison between scheduling strategies provides evidence for the value of intelligent coordination over static sequential processing. The Agent Driven strategy not only achieved higher completion rates (98.2\% vs. 93.1\%) but also demonstrated faster convergence and superior performance across most evaluation dimensions (Figure~\ref{fig:placeholder}). Importantly, this advantage persisted even when the Default Order sequence was designed with clinical considerations in mind, suggesting that dynamic, context-aware task selection offers genuine benefits beyond what can be achieved through careful manual sequencing.

\subsection{Clinical Utility and Physician Perceptions}
\label{subsec:clinical_utility}

The physician evaluation results (Table~\ref{tab:real_world_evaluation}) provide crucial insights into the framework's practical utility. Overall high ratings (all means exceeding 4.45 on a 5-point scale) and low standard deviations (all below 0.23) suggest that the generated medical documentation meets clinical standards and that physicians consistently recognize its quality. The particularly high score for Past History (4.69) indicates that the framework's structured approach to information collection is especially effective for systematically gathering historical medical information—a domain where completeness and accuracy are paramount.

The representative case analysis (Section~\ref{subsec:case_analysis}) illustrates the framework's capability to conduct physician-like structured inquiry. The system demonstrated appropriate clinical questioning patterns: beginning with open-ended symptom exploration, progressing to specific characterization of pain nature and timing, screening for associated symptoms, and systematically collecting past medical history. This progression mirrors established clinical history-taking protocols \cite{gawlik2024historytaking}, suggesting that the multi-agent coordination successfully operationalizes clinical reasoning patterns.

However, we note that the automated Evaluator scores and physician ratings, while both positive, assess different aspects of performance. The Evaluator focuses on predefined dimensions of dialogue quality and content similarity, while physicians bring holistic clinical judgment that may weight factors differently. The general alignment between these assessment modalities provides complementary validation, though the physician evaluation ultimately represents the more clinically meaningful benchmark.

\subsection{Comparison with Prior Work}
\label{subsec:comparison}

Our framework advances the field of intelligent pre-consultation systems in several important respects. Early conversational systems for patient information collection \cite{bickmore2015consent} established basic dialogue capabilities but lacked the sophisticated coordination mechanisms necessary for comprehensive medical history taking. Subsequent work incorporating retrieval-augmented mechanisms \cite{lewis2021retrieval} and reinforcement learning-based multi-agent systems \cite{wang2021collaborative} improved adaptability but remained fundamentally reactive—responding to patient inputs rather than proactively guiding structured inquiry.

Recent LLM-powered approaches, including role-playing agents \cite{tang2024medagents} and modular specialized agent systems \cite{wang2025consultation}, have enhanced clinical fidelity but continue to face challenges with extended dialogue management and global consistency. Our hierarchical task decomposition addresses these limitations by explicitly modeling the pre-consultation workflow as a structured optimization problem with 13 domain-specific subtasks across four main task groups. This formalization enables the Monitor agent to assess subtask completion using clinically-informed criteria and the Controller agent to orchestrate inquiry sequences that balance comprehensiveness with efficiency.

The three core innovations of our framework—dynamic subtask completion assessment, adaptive prompt generation, and hierarchical task management—collectively address the limitations identified in existing systems. The dynamic assessment mechanism (Equation 5) prevents both over-collection (redundant questioning) and under-collection (missing critical information) through threshold-based control. The adaptive prompt generation ensures that follow-up questions remain contextually appropriate and clinically relevant as dialogue progresses. The hierarchical task management enables the system to navigate between macroscopic diagnostic progression and microscopic symptom detail collection—a capability absent from systems employing flat task structures or rigid questioning sequences.

\subsection{Implications and Significance}
\label{subsec:implications}

This research suggests that hierarchical multi-agent frameworks offer a viable approach to addressing systemic healthcare challenges. The documented consultation time constraints—as brief as 4.3 minutes in some settings \cite{wang2022consultation}—create fundamental barriers to diagnostic quality and comprehensive patient care. By automating structured pre-consultation information gathering, systems like ours could enhance diagnostic efficiency, reduce physician cognitive load, and improve documentation completeness without merely redistributing workload to other clinical staff \cite{zhakhina2023pre,DeGroot2022doc}.

The framework's model-agnostic nature has important practical implications. Healthcare institutions vary substantially in their computational resources and data governance requirements. Our demonstration that the hierarchical architecture elicits effective performance across models of different scales and architectures suggests deployment flexibility—institutions could select foundation models based on local constraints while maintaining consistent system behavior through the agent coordination layer.

The success of the Agent Driven scheduling strategy over clinically-designed sequential ordering has broader implications for medical AI system design. Rather than attempting to manually encode optimal clinical workflows, intelligent coordination mechanisms can dynamically adapt to patient-specific information patterns and conversation trajectories. This adaptive capability may be particularly valuable given the inherent variability in patient presentations and the difficulty of anticipating all possible consultation pathways.

\subsection{Limitations}
\label{subsec:limitations}

This study has several important limitations that should be considered when interpreting the findings.

First, our evaluation was conducted using a dataset constructed from a single source (iiyi.com), which, despite encompassing records from diverse hospitals across China, may not fully represent the heterogeneity of patient presentations in other healthcare contexts or geographic regions. The temporal constraint (records from the past five years) ensures contemporary relevance but limits assessment of system performance on historical or unusual presentations.

Second, the framework was evaluated exclusively on Chinese-language medical consultations. While we demonstrated cross-model generalizability, the linguistic and cultural specificity of medical communication patterns may affect performance in other language contexts. The observed efficiency advantage of Qwen3-8B, potentially attributable to stronger Chinese language capabilities, underscores this language-dependent dimension.

Third, although the physician evaluation involved 18 licensed practitioners assessing 40 samples with 9 evaluations per sample, the evaluator pool represents a convenience sample. The relatively narrow standard deviations in physician ratings, while suggesting consistency, may also reflect shared training backgrounds or institutional practices that limit generalizability to broader clinical populations.

Fourth, the current evaluation employed simulated patient interactions through the Virtual Patient agent rather than real patient dialogues. While the Virtual Patient generates responses based on validated medical records, actual patient communication exhibits greater variability, including incomplete responses, tangential information, and emotional factors that may challenge system performance in ways not captured by our evaluation.

Fifth, the 0.85 threshold for subtask completion (Equation 5) was established based on preliminary experimentation and clinical consultation but was not systematically optimized. Different threshold values may yield different balances between consultation efficiency and information completeness, and optimal thresholds may vary across clinical domains or patient populations.

Sixth, safety mechanisms in the current framework rely primarily on prompt-level constraints rather than architectural enforcement. The Recipient and Inquirer agents were explicitly instructed to avoid fabricating medical information and were provided with few-shot examples demonstrating appropriate behaviors. While physician evaluation yielded consistently high scores with low inter-rater variability, suggesting that clinically significant errors were rare in our test cases, we did not conduct systematic adversarial testing or implement dedicated hallucination detection mechanisms. Furthermore, no explicit validation layer exists to intercept potentially inappropriate inquiries before they reach patients. The anonymous nature of our physician evaluation, while ensuring rating fairness, also precluded specialty-stratified analysis of safety perceptions—an important consideration given that physicians in high-acuity specialties may apply more stringent standards. These limitations underscore that the current framework does not guarantee error-free performance, and robust safety validation with architectural safeguards would be essential before clinical deployment.

Finally, we did not conduct direct comparative evaluation against existing pre-consultation systems or human clinician performance on identical cases. While our physician ratings suggest clinical acceptability, quantifying the framework's relative advantage over current practice requires controlled comparative studies.

\subsection{Future Work}
\label{subsec:future_work}

Several directions merit investigation to address current limitations. Evaluating the framework across multiple languages and healthcare systems is essential to assess generalizability beyond Chinese-language consultations.  Prospective validation with real patient interactions would test system robustness to the variability and unpredictability absent in simulated dialogues. Systematic optimization of framework parameters, including 
adaptive completion thresholds that adjust based on clinical context, could enhance performance. Developing architectural safety mechanisms—such as output validation layers, uncertainty quantification, and clinician-in-the-loop checkpoints—represents a critical priority before clinical deployment. Comparative studies benchmarking against existing systems and human clinician performance would provide rigorous evidence of effectiveness. Finally, specialty-specific adaptations may address the performance variability observed across medical departments, particularly for diagnostically challenging domains like psychiatry.

\section{Conclusion}
\label{sec:conclusion}

Effective pre-consultation remains a critical challenge in healthcare systems facing escalating patient volumes and constrained consultation times. This study demonstrated the feasibility of a hierarchical multi-agent framework that transforms passive medical AI systems into proactive inquiry agents through autonomous task orchestration and intelligent coordination. Our framework achieved progressive improvement in triage accuracy, strong cross-model generalizability without task-specific fine-tuning, and high physician ratings for generated medical documentation. The observed divergence in performance across medical specialties and the reliance on simulated patient interactions underscore the need for domain-specific adaptation and prospective validation with real patients. While acknowledging these limitations, our findings suggest that hierarchical task decomposition combined with intelligent agent coordination represents a promising methodological approach for enhancing clinical AI explainability and pre-consultation efficiency, warranting further investigation across diverse healthcare settings.

\section*{Declarations}

\subsection*{Ethics approval and consent to participate}
Ethical review and approval were not required for this study in accordance with local legislation and institutional requirements. This study utilized de-identified electronic health records (EHRs) that are publicly accessible
for viewing on the iiyi.com platform. The data collection process complied with the platform's public accessibility protocols, and no robots.txt or terms of service prohibiting data access were identified. Patient consent was not required as all data were anonymized and obtained from the public domain.

\subsection*{Consent for publication}
Not applicable.

\subsection*{Availability of data and materials}
The source code for the hierarchical multi-agent framework is publicly available at:
\url{https://github.com/SCUT-HCC/MedSynthAI}.

The data used in this study were derived from publicly accessible, de-identified electronic health records
provided by the iiyi.com platform (\url{https://bingli.iiyi.com/}). The raw records were collected via web
crawling and subsequently processed, filtered, and annotated to construct the final dataset used for
analysis.

Due to data ownership considerations and the terms of use of the source platform, the original and
processed datasets cannot be publicly redistributed. However, the de-identified dataset used in this
study may be made available by the corresponding author upon reasonable request, subject to approval
and compliance with the data provider’s terms of use. Detailed descriptions of the data collection and
preprocessing procedures are provided in the manuscript to support reproducibility.

\subsection*{Competing interests}
The authors declare that they have no competing interests.

\subsection*{Ethics approval and consent to participate}

This study was reviewed by the Ethics Review Committee of South China University of Technology and was determined to be exempt from ethical review. The exemption was granted in accordance with relevant regulations, as the study exclusively utilized data collected from publicly accessible websites and public databases. All data used in this research were fully anonymized, meaning that any personal information had been de-identified prior to analysis, and no information could be used to identify specific individuals. As a result, the study does not involve human participants or identifiable private information, and informed consent was therefore not required. This study was conducted in accordance with the principles of the Declaration of Helsinki.

\subsection*{Authors' contributions}
ZJ obtained funding and supervised the study. CY conceived the study, designed the experiments, and drafted the manuscript. YH and ZS constructed the multi-agent system architecture. HC and DM were responsible for data collection and cleaning. ZL provided professional medical advice and conducted the clinical evaluation. NC created the figures and visualizations. All authors read and approved the final manuscript.

\subsection*{Acknowledgments}
The authors gratefully acknowledge the financial support provided by the Guangdong Provincial Department of Science and Technology through Grant 2023CX10X070, the Guangdong Provincial Key Laboratory of Human Digital Twin through Grant 2022B1212010004, the Guangzhou Basic Research Program through Grant SL2023A04J00930, and the Shenzhen Holdfound Foundation Endowed Professorship.


\bibliography{sn-bibliography}

\end{document}